\documentclass[runningheads]{llncs}

\usepackage{wrapfig}
\usepackage{graphics}
\usepackage{threeparttable}
\usepackage{amsfonts}
\usepackage{makecell}
\usepackage{nicefrac}
\usepackage{xcolor}
\usepackage{tabularx}
\usepackage{xspace}
\usepackage{amsmath}
\usepackage{graphicx} 
\usepackage{float} 
\usepackage{subfigure} 
\usepackage{threeparttable}
\usepackage{bbding}

\begin{document}
\title{OneSeg: Self-learning and One-shot Learning based Single-slice Annotation for 3D Medical Image Segmentation}


%
\titlerunning{Single-slice Annotation for 3D Medical Image Segmentation}
\author{Yixuan Wu\inst{1} \and 
Bo Zheng\inst{2} \and 
Jintai Chen\inst{3} \and 
Danny Chen\inst{4} \and 
Jian Wu\inst{5}\inst{(}\Envelope\inst{)} } 
\authorrunning{Y. Wu et al.}
%
\institute{School of Medicine, Zhejiang University, China \and
Polytechnic Institute, Zhejiang University, China\and
College of Computer Science and Technology, Zhejiang University, China\and
Department of Computer Science and Engineering, University of Notre Dame, USA\and
Second Affiliated Hospital School of Medicine, School of Public Health, and Institute of Wenzhou, Zhejiang University, China}
\maketitle              
\begin{abstract}
As deep learning methods continue to improve medical image segmentation performance, data annotation is still a big bottleneck due to the labor-intensive and time-consuming burden on medical experts, especially for 3D images. To significantly reduce annotation efforts while attaining competitive segmentation accuracy, we propose a self-learning and one-shot learning based framework for 3D medical image segmentation by annotating only one slice of each 3D image. Our approach takes two steps: (1) self-learning of a reconstruction network to learn semantic correspondence among 2D slices within 3D images, and (2) representative selection of single slices for one-shot manual annotation and propagating the annotated data with the well-trained reconstruction network. Extensive experiments verify that our new framework achieves comparable performance with less than $1\%$ annotated data compared with fully supervised methods and generalizes well on several out-of-distribution testing sets.
 
\end{abstract}
\section{Introduction}
Recent development of deep learning (DL) methods has revolutionized the medical image segmentation landscape and achieved remarkable successes~\cite{fcnn,unet,3dunet,vnet,deeplab,deeplab0,deeplab2,nnunet,dformer,nnformer}. However, such DL-based segmentation models commonly use large amounts of fine-annotated data in training and the manual annotation process is very time-consuming. Since medical experts, the main annotation providers, are normally busy in daily clinical work, it is highly desirable to develop effective DL segmentation methods with limited annotation, especially for 3D images.

There are two main types of known approaches for alleviating the annotation burden. One type is to explore the potential of non-annotated data~\cite{self1,self2,self3} and use weak-annotation strategies (e.g., using rough annotation as supervision, such as sparse slice labelling~\cite{sparsimi2,rs}, bounding box-level labelling~\cite{bb1,bb2,bb3}, and image-level labelling~\cite{img1,img2,img3}, for medical image segmentation). However, empirical evidences~\cite{ra} suggested that such approaches typically resulted in sub-optimal performance.
 
The second type seeks to annotate only ``worthy'' samples that help improve the final segmentation accuracy, often using active learning methods~\cite{2017mic,2019mic1,2020mic,2021mic}, which iteratively conduct two steps: (i) a model selects valuable samples from the unlabeled set; (ii) experts annotate the selected samples. However, such a process implies that experts should be readily available for queries in each round, and that the active learning process needs to be suspended until queried samples are annotated. To address this, some studies~\cite{ra,rs} managed to avoid human-machine iterations by only selecting representative samples for annotation, but still needed to annotate a considerable amount of samples. 
Inspired by~\cite{mast,track,sli2vol,corrflow,corr2}, the pixel correspondence can be learned in a self-learning manner, and previous work~\cite{mae,mae1,mae2} also verified that a high proportion of pixels in an image (e.g., $80\%$) is redundant and most pixels can be reconstructed from other pixels with self-learning. Base on these, our work investigates to annotate only a single slice of each 3D image and reconstruct full annotation of 3D images using self-learning. 

To this end, in this paper, we propose a self-learning and one-shot learning based framework to predict annotation of any 3D image $X$ if labeled annotation of one selected representative slice in $X$ is provided. 
Note that the word ``one-shot'' in this paper means conducting specific processes only once (i.e., the process of manual annotation and representative slices selection in our framework). 
In general, we train a reconstruction network to learn semantic correspondence among 2D slices of 3D images by self-learning, which is then used to reconstruct and propagate the slice annotation. 
Specifically, in order to learn a effective and robust correspondence, a \textit{screening module} (Sec.~\ref{sec:screening}) is proposed to build a \textit{representative set} in one-shot for self-training and select a most representative single slice of each 3D image for one-shot manual annotation in testing. 
Meanwhile, we design an \textit{information bottleneck} (Sec.~\ref{sec:feature}) to force the model to focus more on high-level semantic features of the images, further contributing to segmentation accuracy. 
To reduce error accumulation when propagating annotation in testing, two training strategies called \textit{scheduled sampling} and \textit{cycle consistency} (Sec.~\ref{sec:loss}) are adopted to enhance robustness of the reconstruction network.

\textbf{Contributions.} (1) We employ self-learning to capture semantic correspondence between the selected \textit{representative slice-pair}, which facilitates the propagation of slice annotation provided by experts. (2) By sharing semantics, one single \textit{representative slice} per 3D image is selected for one-shot manual annotation to address inter-slice redundancy, thus reducing manual annotation efforts. (3) Experiments show that our method performs well with less than 1\% slices manually annotated and generalizes promisingly on several out-of-distribution testing sets.

\subsection{Problem Formulation}
In clinical scenarios, a certain amount of unannotated volumes (3D images) of patients is often available, regarded as a training set, $X_{tr}=\{X_{1},\ldots,X_{N}\}$, containing enormous semantics to be utilized (though unannotated). Meanwhile, raw volumes of new patients are constantly produced, regarded as an testing set, $X_{te}=\{X_{1}^\prime,\ldots,X_{M}^\prime\}$, needed to be segmented for clinical diagnosis and analysis. Each volume has $D$ slices ($D$ could be different for volumes), $X_{i}=\{S_{i1},\ldots,S_{iD}\}$, with $S_{id} \in \mathbb{R}^{H\times W}$. In this setting, we propose a novel framework to attain full annotation, i.e., $\{\hat{A}^{\prime}_{i1},\ldots,\hat{A}^{\prime}_{iD}\}$ of any volume $X^{\prime}_{i}$ in the testing set with only one slice's ground truth annotation $A_{id}^\prime$ provided.

\subsection{Pipeline}
\begin{figure}[t] 
\centering 
\includegraphics[width=12cm,height=6.22cm]{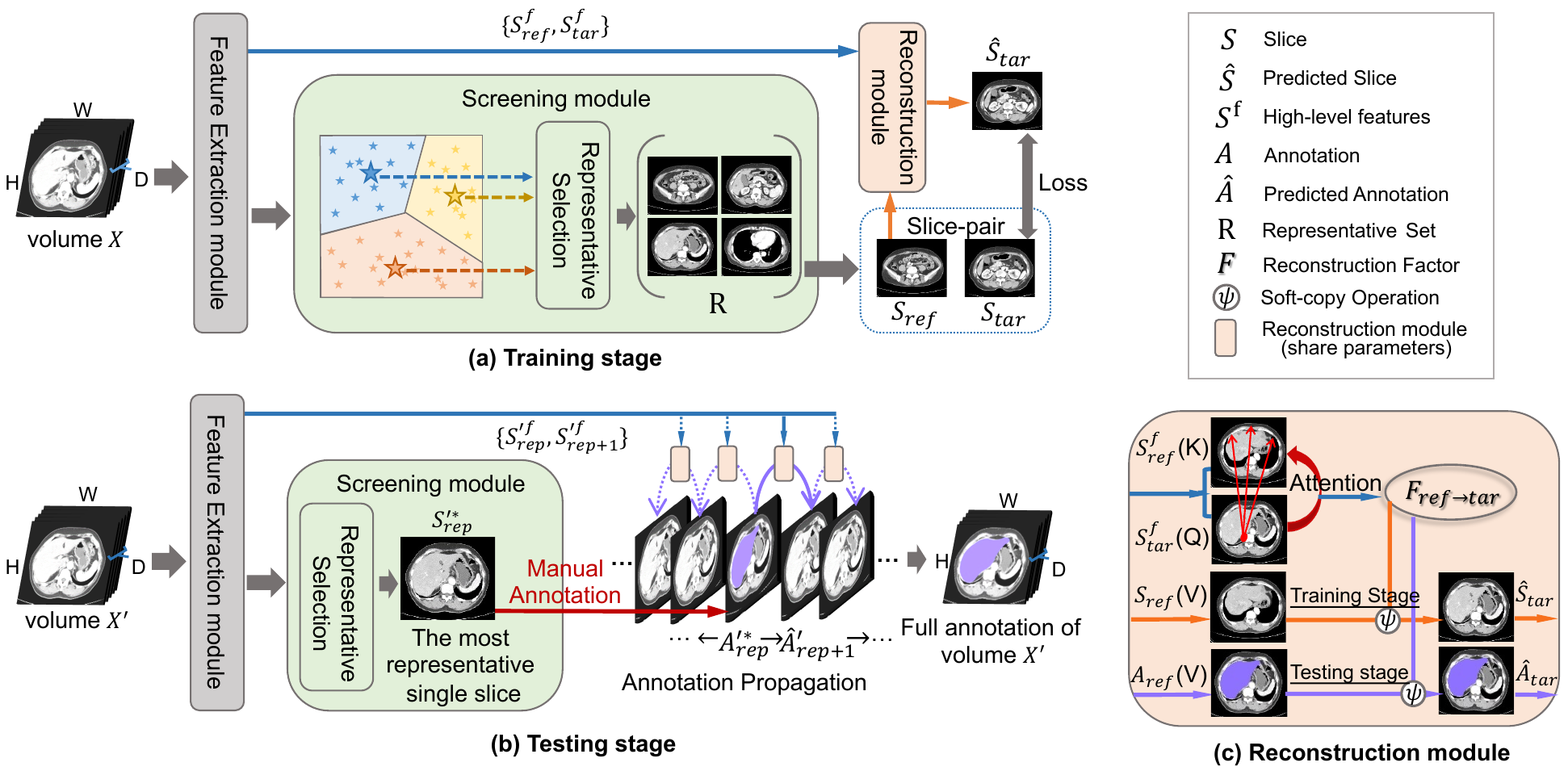}
\caption{The pipeline of our framework. 
(a) In training, the \textit{representative set $R$} is constructed, from which several pairs of slices (i.e., \textit{representative slice-pair}) are then sampled for reconstructing the one from another in a self-learning manner.
(b) In testing, only a single slice is screened out for manual annotation, and then the annotation of all remaining slices are reconstructed.
(c) Specification of \textit{Reconstruction module}. 
} 
\label{fig:pip} 
\end{figure}

The main idea of our method is to learn pixel correspondence among slices, and use it to reconstruct the annotation of target slices.
To this end, during training, a proxy task is proposed as reconstructing a target slice $S_{tar}$ based on a reference slice $S_{ref}$ by self-learning within a series of selected \textit{representative slice-pair} $\{S_{ref},S_{tar}\}$. \\
\noindent \textbf{Training Stage.} 
As illustrated in Fig.~\ref{fig:pip}(a), with our proposed \textit{feature extraction module} (Sec.~\ref{sec:feature}) and \textit{screening module} (Sec.~\ref{sec:screening}), a series of \textit{representative slice-pairs} is selected from the unannotated training volumes in one-shot (for simplicity, Fig.~\ref{fig:pip}(a) shows only one input volume and one \textit{representative slice-pair}). Then, a reference slice $S_{ref}$ and corresponding high-level features $\{S_{ref}^{f},S_{tar}^{f}\}$ obtained by the \textit{feature extraction module} are fed to the \textit{reconstruction module} (Sec.~\ref{sec:reconstruction}) to reconstruct the target slice $\hat{S}_{tar}$. \\
\noindent \textbf{Testing Stage.} 
As shown in Fig.~\ref{fig:pip}(b), for any volume $X^\prime$, its most representative single slice $S^{\prime*}_{rep}$ is screened out by the well-trained \textit{feature extraction module} and \textit{screening module} in one-shot, and manual annotation is required only for this single slice. Then, the obtained ground truth annotation $A^{\prime*}_{rep}$ and its corresponding high-level features $\{S_{rep}^{\prime f},S_{rep+1}^{\prime f}\}$ are computed by the \textit{reconstruction module} in which an adjacent slice's annotation  $\hat{A}^{\prime}_{rep+1}$ is reconstructed. In this manner, annotation will be propagated slice-by-slice, until all slices in this volume are (computationally) annotated, i.e., $\hat{A}_{1}^\prime,\ldots,A_{rep}^{\prime*},\ldots,\hat{A}_{D}^\prime$.

\subsection{Feature Extraction Module}\label{sec:feature}
To obtain feature representations of slices for the \textit{screening} and \textit{reconstructing} processes, we design the \textit{feature extraction module} to extract high-level features, as:
\begin{gather}
    S_{i}^{f}=\Phi(g(S_{i}),\theta),
\end{gather}
where $S_{i}^{f}$ denotes the extracted high-level features of a slice $S_{i}$, $\Phi(\cdot,\theta)$ refers to the ResNet-18~\cite{resnet} feature encoder, and $g(\cdot)$ denotes our \textit{information bottleneck}. The \textit{information bottleneck} aims to force the model to focus more on high-level semantic features instead of simple pixel intensities when learning correspondences between slices. We adopt the Gabor filter~\cite{gabor} as the \textit{information bottleneck} to emphasize the edge and texture features. For each slice $S_{i}\in \mathbb{R}^{H\times W}$, after the Gabor feature extraction, we obtain $S$ different scales and $O$ different orientations of filtered features, which are pixel-wise concatenated along the channel dimension to obtain $g(S_{i})\in\mathbb{R}^{H\times W\times S\times O}$. Then, after processed by ResNet-18, high-level features $S_{i}^{f}\in\mathbb{R}^{H\times W\times C}$ are obtained. In our implementation, $S,O$, and $C$ are set to $4,8$, and $256$, respectively.

\subsection{Screening Module}\label{sec:screening}
The key challenge to our method is how to select suitable slice-pairs for self-learning and how to screen out the most ``worthy'' single slice to annotate in testing. To this end, we propose the \textit{screening module} to select suitable samples in one-shot, as shown in Fig.~\ref{fig:pip}. \\
\noindent \textbf{Training Stage.}
For a medical volume, its adjacent slices are often similar but all the slices may be divided into several consecutive sub-sequences based on anatomical structures. To deal with redundancy between adjacent slices and learn the correspondence between slices with diverse anatomical structures, the \textit{screening module} selects a \textit{representative set $R$} from all the slices of each training volume $X$. 
First, $K$-Means clustering~\cite{kmeans} is performed on all the slices based on their high-level features $\{S_{1}^{f},\ldots,S_{D}^{f}\}$, where $K=D/I$ and $I$ are randomly selected from $\{2,3,5\}$. After clustering, each cluster $C_{k}\enspace(k=1,\ldots,K)$ contains $N_{k}$ slices, i.e., $C_{k}=\{S_{kj}\enspace|\enspace j=1,\ldots,N_{k}\}$. 
Second, the \textit{representative selection} is conducted where the most representative slice $S_{c_{k}}^{*}$ from each cluster $C_{k}$ is selected to form the \textit{representative set $R$} based on max cosine similarity, as:
\begin{gather}
    Sim(S_{i},S_{j})=Cosine\_similarity(S_{i}^{f},S_{j}^{f}), \label{eq:ms1}\\
    RepScore(S_{i},C_{k})=\sum_{S_{j}\in C_{k}} Sim(S_{i},S_{j}),\label{eq:ms2}\\
    S_{c_{k}}^{*}=\arg\max_{S_{i}\in C_{k}}(RepScore(S_{i},C_{k})),\label{eq:ms3}
\end{gather}
where $ Sim(S_{i},S_{j})$ denotes the cosine similarity between slices $S_{i}$ and $S_{j}$, $RepScore$ denotes the representative score of slice $S_{i}$ for cluster $C_{k}$, and $S_{c_{k}}^{*}$ is the most representative slice for cluster $C_{k}$. Third, all the selected slices from the clusters form the \textit{representative set} $R=\{S_{c_{1}}^{*},\ldots,S_{c_{K}}^{*}\}$, and \textit{representative slice-pairs} are selected from $R$ one-by-one for subsequent self-learning, i.e., $\{S_{c_{1}}^{*},S_{c_{2}}^{*}\},\{S_{c_{2}}^{*},S_{c_{3}}^{*}\},\ldots$. Note that the above process is conducted on one volume, and all the volumes in the training set are processed in the same way. \\
\noindent \textbf{Testing Stage.} 
For each testing volume with $D$ slices, i.e., $X^\prime=\{S^\prime_{1},\ldots,S^\prime_{D}\}$, the \textit{representative selection} is conducted in one-shot to screen out the most representative single slice $S_{rep}^{\prime*}$, as:
\begin{gather}
    Sim(S_{i}^\prime,S_{j}^\prime)=Cosine\_similarity(S_{i}^{\prime{f}},S_{j}^{\prime{f}}), \label{eq:ms4}\\
    RepScore(S_{i}^\prime,X^\prime)=\sum_{S_{j}^\prime\in X^\prime} Sim(S_{i}^\prime,S_{j}^\prime),\label{eq:ms5}\\
    S_{rep}^{\prime*}=\arg\max_{S_{i}^\prime\in X^\prime}(RepScore(S_{i}^\prime,X^\prime)). \label{eq:ms6}
\end{gather}
 
\subsection{Reconstruction Module}\label{sec:reconstruction}
The \textit{reconstruction module} aims to reconstruct a target slice $S_{tar}$ by linearly combining pixels from a reference slice $S_{ref}$, with weights measuring the strengths of correspondences between pixels. In detail, in order to reconstruct pixel $i$ in the target slice, an attention mechanism~\cite{track} is applied to measure the similarity between pixel $i$ in the target slice and a related pixel $j$ in the reference slice, denoted as \textit{reconstruction factor} $F_{ref\rightarrow{tar}}^{ji}$, which is used to re-weight pixel $j$ to reconstruct pixel $i$. All the pixels in the target slice are reconstructed in this manner to obtain the reconstructed target slice $\hat{S}_{tar}$:
\begin{gather}
    F_{ref\rightarrow{tar}}^{ji}=\frac{\exp\langle Q_{tar}^{i},K_{ref}^{j}\rangle}{\sum_{p}\exp\langle Q_{tar}^{i},K_{ref}^{p}\rangle},\\
    \hat{S}_{tar}=
    \Psi(F_{ref\rightarrow{tar}},V_{ref})=
    \sum_{i}\sum_{j}F_{ref\rightarrow{tar}}^{ij}V_{ref}^{j},
\end{gather}
where $\langle \cdot,\cdot \rangle$ denotes the \textit{dot product} between two vectors, query ($Q$) and key ($K$) are high-level features computed by the \textit{feature extraction module} (i.e., $Q_{tar}=\Phi(g(S_{tar}),\theta),K_{ref}=\Phi(g(S_{ref}),\theta)$), $V$ is a reference slice in training (i.e., $V_{ref}=S_{ref}$) and refers to the reference slice's annotation in testing (i.e., $V_{ref}=A_{ref}$), $i,j$ denote corresponding pixels in the target and reference slices respectively, $p$ denotes a pixel in a square patch of size $P\times P$ surrounding pixel $j$ ($P= 13$ in our implementation), and $\Psi(\cdot,\cdot)$ denotes the soft-copy operation for pixel-wise reconstruction. 
\subsection{Training Strategies}\label{sec:loss}
\noindent\textbf{Scheduled Sampling.}
In our original setting above, a \textbf{ground truth} reference slice is used to reconstruct a target slice in training, while in testing, it is a \textbf{reconstructed} annotation that is used to reconstruct its adjacent annotation iteratively, which may lead to possible error accumulation. To bridge this gap between training and testing, we adopt \textit{scheduled sampling}~\cite{schedule} to replace some ground truth reference slices by reconstructed slices in training. We define the loss as $\mathcal{L}_{sche}$:
\begin{gather}
    \mathcal{L}_{sche}=
    \alpha_{1}\cdot\sum_{i=1}^{n}\Vert S_{i+1}-\Psi(F_{i\rightarrow{i+1}},S_{i})\Vert _{1}
    +
    \alpha_{2}\cdot\sum_{i=1}^{n}\Vert S_{i+1}-\Psi(F_{i\rightarrow{i+1}},\hat{S_{i}})\Vert _{1},
\end{gather}
where $\{S_{i},S_{i+1}\}$ denotes \textit{representative slice-pairs} from \textit{representative set $R$}, and $n$ denotes the number of \textit{representative slice-pairs} for training. $\Psi(\cdot,\cdot)$ is the soft-copy operation for pixel-wise reconstruction, and $F_{i\rightarrow{i+1}}$ denotes \textit{reconstruction factor}. The weight $\alpha_{1}$ starts from a high value of 0.9 in early training stage and is uniformly annealed to 0.5, keeping $\alpha_{1}+\alpha_{2}=1$.\\
\noindent\textbf{Cycle Consistency.}
Normally, in testing, the expert-provided single slice's annotation $A_{i}$ is around the middle of the slices along the depth dimension of a volume, and the annotation is propagated in two directions, i.e., $A_{i}\rightarrow{\hat{A}_{i+1}}$ and $A_{i}\rightarrow{\hat{A}_{i-1}}$. \textit{Cycle consistency}~\cite{cycle} is adopted to enhance the model's robustness for reconstruction in both directions. Specifically, for one \textit{representative slice-pairs}, $\{S_{i},S_{i+1}\}$, $S_{i}$ is used to reconstruct $\hat{S}_{i+1}$ in one direction, i.e., $i\rightarrow{i+1}$, and, in turn, the reconstructed slice $\hat{S}_{i+1}$ is also utilized to reconstruct the initial one $\hat{S}_{i}$ in the other direction, i.e., $i+1\rightarrow{i}$. The loss $\mathcal{L}_{cyc}$ is defined as:
\begin{gather}
    \mathcal{L}_{cyc}=\sum_{i=1}^{n}\Vert S_{i}-\Psi(F_{i+1\rightarrow{i}},\Psi(F_{i\rightarrow{i+1}},S_{i}))
    \Vert_{1}.
\end{gather}

Thus, the overall learning objective loss $\mathcal{L}$ is defined as:
\begin{gather}
    \mathcal{L}=\lambda_{1}\mathcal{L}_{sche}+\lambda_{2}\mathcal{L}_{cyc},
\end{gather}
where the weights $\lambda_{1}$ and $\lambda_{2}$ are set to 0.9 and 0.1, respectively. 

\section{Experiments}
\subsection{Datesets and Metrics}
We conduct experiments on four public benchmark datasets with different imaging modalities: LiTS~\cite{lits}, CHAOS~\cite{chaos}, Sliver07~\cite{sliver}, and Decathlon-Liver~\cite{decathlon}. For CT, our model is trained on LiTS that includes 130 contrast-enhanced abdominal CT scans, and is tested on CHAOS (a CT subset), Sliver07, and Decathlon-Liver which contain 20, 20, and 131 abdominal CT scans, respectively. For MRI, our model is both trained and tested on CHAOS (an MRI subset) that includes 60 training samples and 60 testing samples (20 T1-DUAL in-phase, 20 oppose-phase, and 20 T2-SPIR, respectively). Following the assessment criteria of the CHAOS challenge~\cite{chaos}, we use Dice coefficient (DICE), relative absolute volume difference (RAVD), and average symmetric surface distance (ASSD) as metrics to evaluate
results based on overlapping, volumetric, and spatial differences.
\subsection{Implementation}
Our model is implemented with PyTorch 1.8.0, and all the experiments are conducted on an NVIDIA GeForce RTX 2080 GPU with 11 GB memory. The batch size during training is 2 and during testing is 1. The Adam optimizer~\cite{adam} is used with an initial learning rate of 0.0001, which is halved every epoch. In pre-processing, all slices are resized to $256\times256$. 
\begin{table}[h]
\footnotesize
\centering
\caption{Segmentation performances of different methods on CHAOS.}
\label{tab:acc}
\setlength{\tabcolsep}{0.01pt}
\begin{tabular}{l|ccc|ccc}
\hline
\multicolumn{1}{c|}{Modality (Organ)}                                      & \multicolumn{3}{c|}{MRI (Spleen)} & \multicolumn{3}{c}{CT (Liver)} \\ \hline
\multicolumn{1}{c|}{Method}                                       & DICE$\uparrow$     & RAVD$\downarrow$      & ASSD$\downarrow$     & DICE$\uparrow$    & RAVD$\downarrow$     & ASSD$\downarrow$    \\ \hline
(1) 2D U-Net (one slice annotated) \cite{unet}  & 0.64     & 39.36     & 31.57    & 0.80    & 17.75    & 14.94   \\
(2) 3D U-Net (one slice annotated) \cite{3dunet}  & 0.47     & 37.52     & 34.25    & 0.57    & 19.78    & 17.54   \\
(3) Pseudo annotation \cite{pseudo}                                                        & 0.69     & 22.41     & 14.36    & 0.63    & 22.24    & 25.34   \\
(4) Scribble-level annotation \cite{scribble} & 0.72     & 21.67     & 13.54    & 0.81    & 11.37    & 12.22   \\
(5) Box-level annotation \cite{box}                                                 & 0.73     & 21.25     & 9.98     & 0.79    & 15.79    & 9.97    \\
OneSeg (Ours)                                                                        & \textbf{0.85}     & \textbf{11.58}     & \textbf{4.35}     & \textbf{0.93}    & \textbf{7.48}     & \textbf{3.34}    \\ \hline
(6) 3D U-Net (Fully annotated) \cite{3dunet}                                                  & 0.89     & 9.74      & 4.21     & 0.94    & 13.74    & 3.47    \\ \hline
\end{tabular}
\end{table}
\begin{table}[t]
{\footnotesize
\centering
\caption{Results of two methods when the testing set shifted from the training set.}
\label{tab:gen}
\resizebox{\textwidth}{10mm}{
\footnotesize
\begin{tabular}{c|c|c|c}
\hline
Training set (Organ)                           & \multicolumn{3}{c}{LiTS (Liver)}                     \\ \hline
Testing set (Organ)                            & Sliver07 (Liver) & CHAOS (Liver) & Decathlon (Liver) \\ \hline
Method                                          & DICE $\uparrow$             & DICE $\uparrow$          & DICE $\uparrow$              \\\hline
\multicolumn{1}{l|}{3D U-Net (Fully annotated) \cite{3dunet}} & 0.71             & 0.75          & 0.63              \\
\multicolumn{1}{l|}{OneSeg (Ours)}                       & \textbf{0.92}    & \textbf{0.93} & \textbf{0.89} \\ \hline    
\end{tabular}}
}
\end{table}
\subsection{Compared Methods and Results}
\noindent \textbf{Compared Methods.} To investigate the segmentation accuracy of our method, we compare it with common known methods that use different annotation strategies, including: (1) 2D U-Net~\cite{unet} trained on one annotated slice in each 3D training volume, (2) 3D U-Net~\cite{3dunet} trained on 3D training volumes with one 2D slice annotated and other slices unannotated, (3) 3D CNN~\cite{pseudo} trained on one fully annotated volume and augmented pseudo annotation, (4) 3D CNN~\cite{scribble} trained on scribble annotation where the location of a target organ in all slices in each training volume is specified by a small number of voxels, (5) 3D CNN~\cite{box} trained on bounding-box annotation of all slices in each training volume, (6) 3D U-Net~\cite{3dunet} trained on all fully annotated training volumes (representing the performance of a fully-supervised model). For fair comparison, the annotated slices in (1) and (2) are randomly selected and kept the same. All the compared methods work in the original settings of their papers, which are trained/tested on the CT and MRI subsets of CHAOS, respectively, based on official division~\cite{chaos}.  

\noindent \textbf{Performance.}
From the results in Table~\ref{tab:acc}, (1) and (2) are trained with the same amount of annotation data as our method, but show inferior performance, which suggests that single slice annotation is too sparse for conventional fully supervised models. Besides, compared with other weakly-supervised methods ((3), (4), and (5)), our method achieves superior accuracy by a large margin. Moreover, our method shows close results to (6) while utilizing only less than $1\%$ annotation (i.e., one 2D slice vs.~a 3D volume). Notably, our method even outperforms (6) in the ASSD metric that is for evaluating model performance with boundary and spatial variations. \\
\noindent \textbf{Generalizability.} To explore the generalizability of our method, we compare it with fully supervised 3D U-Net~\cite{3dunet} by testing on several out-of-distribution testing sets.
Specifically, both the methods are trained on LiTS and tested on CHAOS, Sliver07, and Decathlon-Liver, respectively. In table~\ref{tab:gen}, the performances of fully-supervised 3D U-Net drop considerably when shifting to other datasets for testing, even when the target organ and modality remain the same.
In contrast, our method shows promising generalizability.
Medical images are commonly collected from different machines, often leading to domain shift. In such scenarios, the generalizability allows our model to attain good performances on out-of-distribution testing sets, thus achieving robustness in predicting annotation for new images. 

\begin{table}[t]
\footnotesize
\centering
\caption{Ablation study on the effect of each module on CHAOS (a CT subset).}
\label{tab:abl}
\begin{tabular}{l|ccc}
\hline
\multicolumn{1}{c|}{Method} & DICE $\uparrow$                     & RAVD $\downarrow$                     & ASSD $\downarrow$                     \\ \hline
w/o Screening Module         & 0.74                     & 27.54                    & 18.24                    \\
w/o Information Bottleneck   & 0.84                     & 14.12                    & 10.23                    \\
w/o Scheduled Sampling       & 0.80                     & 21.02                    & 12.24                    \\
w/o Cycle Consistency        & 0.89                     & 8.24                     & 7.94                     \\
Full Model, $\lambda_{1}=0.1, \lambda_{2}=0.9$  & 0.83                     & 15.44                     & 11.78                     \\
Full Model, $\lambda_{1}=0.5, \lambda_{2}=0.5$  & 0.87                     & 10.23                     & 9.13                     \\
Full Model, $\lambda_{1}=0.9, \lambda_{2}=0.1$                   & \textbf{0.93} & \textbf{7.48} & \textbf{3.34} \\ \hline
\end{tabular}
\end{table}
\subsection{Ablation Study}
To evaluate the effectiveness of our method, we assess our full model against the absence of each its module on CHAOS (a CT subset). Table~\ref{tab:abl} shows that the \textit{screening module} contributes to the results by 0.19 in DICE, compared to randomly taking a single slice for manual annotation. Besides, without \textit{information bottleneck}, only ResNet-18 is used to extract features, which leads to an accuracy drop of 0.09 in DICE. Moreover, the two training strategies, \textit{scheduled sampling} and \textit{cycle consistency}, also improve the performance by 0.13 and 0.04 in DICE, respectively. Meanwhile, we try different weights of these two strategies and $\lambda_{1}=0.9, \lambda_{2}=0.1$ work well. Thus, it is evident that each our module contributes to the segmentation performance.  

\section{Conclusions}
In this paper, we proposed a new self-learning and one-shot learning based framework for 3D medical image segmentation with only a single slice annotation provided. The single \textit{representative slice} of each 3D image is selected for one-shot manual annotation, thus reducing annotation efforts. Self-learning of a reconstruction network facilitates to match semantic correspondences between the well-constructed \textit{representative slice-pair}.
Extensive experiments showed that with less than $1\%$ annotated data, our method achieves competitive results and promising generalizability. \\


\bibliographystyle{splncs04}
\bibliography{refer}
\end{document}